\documentclass[lettersize,journal,10pt]{IEEEtran}
\usepackage{amsmath,amsfonts}
\usepackage[ruled,linesnumbered]{algorithm2e}
\usepackage{array}
\usepackage[caption=false,font=normalsize,labelfont=sf,textfont=sf]{subfig}
\usepackage{textcomp}
\usepackage{stfloats}
\usepackage{url}
\usepackage{verbatim}
\usepackage{graphicx}
\usepackage{amsthm}
\usepackage{amssymb}
\def\BibTeX{{\rm B\kern-.05em{\sc i\kern-.025em b}\kern-.08em
    T\kern-.1667em\lower.7ex\hbox{E}\kern-.125emX}}
\usepackage{balance}

\newtheorem{definition}{\indent Definition}

\begin{document}
\title{Contextual Bandits with Non-Stationary \\ Correlated Rewards for User Association\\ in MmWave Vehicular Networks}

\author{Xiaoyang He, Xiaoxia Huang,~\IEEEmembership{Senior Member,~IEEE,} Lanhua Li,~\IEEEmembership{Member,~IEEE}}



\maketitle

\begin{abstract}
	Millimeter wave (mmWave) communication has emerged as a propelling technology in vehicular communication.
	Usually, an appropriate decision on user association requires timely channel information between vehicles and base stations (BSs), which is challenging given a fast-fading mmWave vehicular channel.
	In this paper, relying solely on learning transmission rate, we propose a low-complexity semi-distributed contextual correlated upper confidence bound (SD-CC-UCB) algorithm to establish an up-to-date user association without explicit measurement of channel state information (CSI).
	Under a contextual multi-arm bandits framework, SD-CC-UCB learns and predicts the transmission rate given the location and velocity of the vehicle, which can adequately capture the intricate channel condition for a prompt decision on user association.
	Further, {SD-CC-UCB efficiently identifies} the set of candidate BSs which probably support supreme transmission rate by leveraging the correlated distributions of transmission rates on different locations.
	To further refine the learning transmission rate over the link to candidate BSs, each vehicle deploys the Thompson Sampling algorithm by taking the interference among vehicles and handover overhead into consideration.
	Numerical results show that our proposed algorithm achieves {the network throughput} within {100\%-103\%} of a benchmark algorithm which requires perfect instantaneous CSI, demonstrating the effectiveness of SD-CC-UCB in vehicular communications.
\end{abstract}


\section{Introduction}
\IEEEPARstart{C}{apable} of Supporting high transmission rates, massive access, and low latency, mmWave technology is being envisaged for various applications, such as fifth-generation (5G) networks and automated driving technology \cite{9779354}.
In particular, establishing the connection between users and base stations (BSs), known as user association plays a crucial role in networking \cite{7378276}.
A potentially volatile association between the user and the BS can result in diminished network throughput.
{Each user is associated with the BS with the highest Signal to Interference plus Noise Ratio (SINR) in 3GPP \cite{7156092}.}
Users need to measure the SINR to each BS timely before associations, which introduces notable delays and overhead {due to the rapidly changing channel conditions and the densely deployed BSs in} mmWave vehicular networks.
Moreover, the max-SINR strategy results in substantial handovers due to the fluctuations of the mmWave vehicular channel, leading to significant signaling overhead and throughput degradation.

User association is faced with tremendous challenges in mmWave vehicular network, including the fast-fading characteristics of the channel and interference introduced by vehicles.
{The channel conditions between users and BSs are necessary to determine the appropriate associations.}
{However, the channel conditions are time-varying, because  the mmWave signal can hardly penetrate blockages and experiences rapidly changing Line-of-Sight (LOS) and Non-Line-of-Sight (NLOS) propagation due to the short wavelength of mmWave and the high mobility of vehicles.}
Moreover, the rapidly moving vehicles introduce a significant Doppler effect \cite{8594703}, which further exacerbates the instability of channels.
A user may maintain an association with a BS for a very brief period, sometimes as short as 0.75 seconds \cite{7022933}.
{Acquiring instantaneous channel conditions incurs heavy signalling overhead which increases with the density of users, making it less effective in dense mmWave vehicular networks.}
Besides, CSI is insufficient to determine the optimal user association because interference plays a significant role in transmission rates.
The channel with a good CSI may suffer from substantial interference, resulting in a poor transmission rate.
{For example, when a traffic jam occurs, some BSs close to jamming locations, namely ``hot spots" will be connected with many jammed vehicles.}
The excessive transmission demand introduces intensive interference among jammed vehicles towards the hot spots, which must be captured in user association and resource allocation.

{It is essential to develop a user association algorithm which optimizes the network throughput in mmWave vehicular networks.
	This problem is conventionally addressed as an integer non-convex programming, typically known to be NP-hard \cite{6497017}.}
The problem is even more severe in the mmWave vehicular network because the user association needs to be invoked frequently as the channel conditions dramatically change.
Prior work has attempted to reduce the complexity using the Lagrangian duality theory or a heuristic algorithm \cite{6497017,6774981,8677293,7961156,9417380}.
For instance, the worst connection swapping (WCS) algorithm \cite{8677293} identifies the user with the poorest connection and switches from the currently associated BS to the optimal BS in each iteration.
However, WCS is a centralized iterative algorithm and requires CSI to estimate the throughput considering interference.
The time complexity required for CSI calculation is proportional to the number of users and BSs \cite{9732214}.
The frequent channel measurement or estimation is too costly {in mmWave vehicular networks.}
\begin{figure}[htbp]
	\centering\includegraphics[scale=0.3]{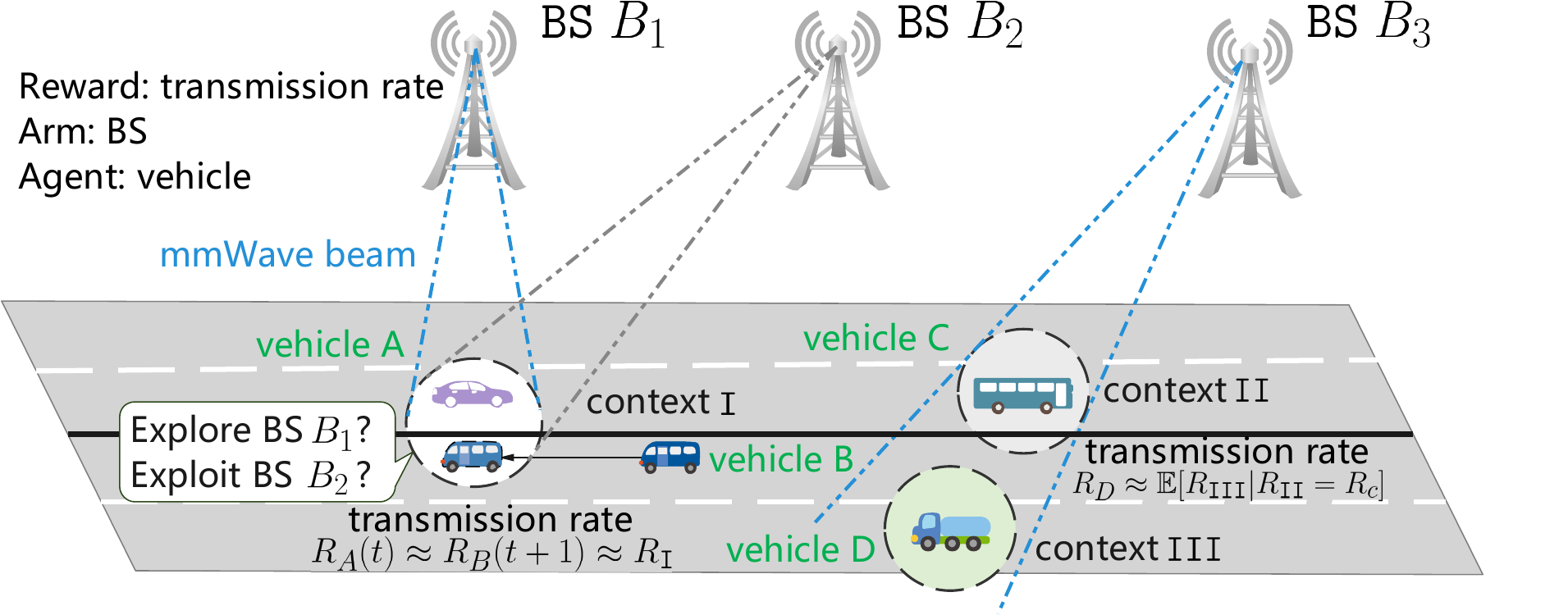}
	\caption{A multi-user multi-BSs vehicular network, where vehicles within the same context experience a comparable channel condition.}
	\label{Fig_introduction}
\end{figure}

With affordable computation complexity and moderate requirements on channel conditions, online learning is a promising technique for efficient user association, which offers a low-complexity alternative to the non-convex optimization problem \cite{8472783,9286435,9789983,9322601}.
The multi-armed bandits (MAB) framework is a typical model applied in online learning.
In the MAB framework, BSs and vehicles can be considered as arms and agents, respectively.
A learning agent takes actions (communicates to a BS), and interacts with the environment to receive rewards (instantaneous transmission rate).
The agent aims to achieve a trade-off between acquiring new information about reward distributions through exploration, and exploitation based on available information to maximize the cumulative reward \cite{auer2002finite}.
As shown in Fig.~\ref{Fig_introduction}, vehicle B pulls the arm with the highest reward, which is BS $B_2$ at period $t$.
As it travels to a new location in the next period, the vehicle has to decide whether to maintain its transmission with BS $B_2$ (exploitation) or to communicate with another BS $B_1$ (exploration).
The MAB framework is highly adaptable and capable of learning from the action and environment immediately, allowing it to make proper decisions in response to the dynamic environment.

However, existing MAB algorithms cannot well suit the unique characteristics and challenges in mmWave vehicular networks for the following reasons.
{First, some classical MAB algorithms, such as the upper confidence bound (UCB), rely on the assumption that the reward distribution is stationary \cite{orabona2019modern}.}
Unfortunately, this underlying assumption is violated by the fast-fading mmWave vehicular channel.
{The channel gain varies with the vehicle's velocity, location, and blockage which determine the Doppler spread, LOS, and NLOS links between the vehicle and a BS.}
Therefore, the algorithms relying on the assumption of stationary reward distribution may perform poorly.
{As an extension of MAB, contextual MAB is particularly useful for action decisions given non-stationary reward distributions \cite{9185782}.
	In contextual bandits, the reward distribution within an individual context can be considered stationary.}
{
	The contextual MAB framework has been applied to capture the non-stationary reward distributions induced by time-varying traffic congestions and blockages, incorporating the arrival directions and velocities of vehicles as context information  \cite{8472783,li2020smart}.
	However, it is still difficult to address the non-stationary reward distributions posed by fast-fading channels and the severe Doppler effect in mmWave vehicular networks.
	Moreover, due to the expansive coverage area of the vehicular network, the context space tends to be tremendous.
	It is challenging to effectively learn with insufficient samples for each context in these contextual MAB-based algorithms.}

In this paper, we propose a semi-distributed contextual correlated upper confidence bound (SD-CC-UCB) algorithm that identifies favorable BSs and explores arms efficiently by leveraging the correlation between contextual rewards.
We first devise a centralized contextual correlated UCB (CC-UCB) algorithm to learn the channel condition embodied with the transmission rate given locations and velocities of vehicles with a reasonable number of rate measurements.
The CC-UCB algorithm incorporates the vehicle's location and velocity as context information.
Within each context, the location and velocity are comparable, leading to comparable channel geometrical characteristics and Doppler spread.
For example, when vehicles A and B are located in the dashed circle (context) shown in Fig.~\ref{Fig_introduction}, they experience comparable channel conditions.
Although vehicles arrive at context I at different time, they encounter similar LOS and NLOS links.
{Unlike solely updating the reward for the corresponding context \cite{8472783,li2020smart}, our proposed CC-UCB algorithm innovatively utilizes each sample to learn about the reward distributions for not only the corresponding context but also other contexts.}
For example, vehicle C within context II communicates with BS $B_3$ and receives a reward $R_C$, as shown in Fig. \ref{Fig_introduction}.
{CC-UCB not only updates the empirical reward of context II but also can obtain some priori knowledge to learn about the reward distribution of context III.}
This priori knowledge helps vehicle D on context III to identify whether BS $B_3$ is worth exploring so that vehicles can explore more efficiently.
We refer to the BSs worth exploring as the competitive BSs.
Further, considering the interference from other vehicles and handover cost, the problem is modeled as multi-agent multi-armed bandits (MA-MAB) and solved with a distributed Thompson Sampling (TS) algorithm at low computation cost.

The main contributions of this paper are summarized as follows.
\begin{itemize}
	\item
	{
		The fast-fading channels and the severe Doppler effect in mmWave vehicular networks result in non-stationary reward distributions.
		Within a contextual MAB framework, the vehicle's location and velocity are incorporated as context information.
		Then, the non-stationary reward is converted into stationary rewards within a context during a short horizon.}
	
	\item
	The extensive coverage area of the vehicular network generates a substantial context space, so the current contextual MAB algorithm converges slowly.
	{We discover that a sample's reward can be considered as the upper bound of the rewards for other contexts.
		The proposed CC-UCB algorithm can eliminate the arms with low upper bounds of rewards, thereby ruling out unnecessary explorations and achieving quick convergence.}
	\item
	To further enhance the accuracy of the estimated transmission rate, each vehicle deploys the TS algorithm to determine the associated BS within the competitive BSs generated by CC-UCB.
	Vehicles can identify and avoid choosing the BS with a bad channel condition or the BS with a good channel condition but significant interference by considering the impact of interference introduced by other vehicles.
	\item
	We derive an upper bound on the expected cumulative regret of the SD-CC-UCB algorithm and {demonstrate that our proposed algorithms can achieve a low regret in mmWave vehicular networks attributed to the identified competitive set.}
	Extensive simulations have demonstrated the effectiveness of our proposed SD-CC-UCB algorithm, achieving the network throughput within 100\%-103\% of the centralized offline WCS benchmark algorithm, which requires costly instantaneous CSI between all vehicles and BSs \cite{8677293}.
\end{itemize}

The rest of this paper is organized as follows. 
The related work is introduced in section II.
In Section III, we provide the mobility model of vehicles and the channel model, along with the problem formulation and the overview of our proposed algorithm.
Then, the CC-UCB and SD-CC-UCB algorithms are proposed in Sections IV and V, respectively.
In Section VI, simulation results validate the effectiveness of our algorithms. 
Finally, Section VII concludes the paper.

\section{Related work}
User association in vehicular networks is crucial for managing mobility, balancing network load, ensuring quality of service (QoS), optimizing resource usage, improving network reliability, and promoting energy efficiency.
Effective user association strategies are indispensable to ensuring the seamless functionality of vehicular communication systems.
{Nonetheless, user association is a computation intensive problem belonging to the integer non-convex programming.}
In response to this challenge, various offline algorithms have been proposed to attain a near-optimal solution for the NP-hard problem within polynomial time constraints {\cite{6497017,6774981}}.
In contrast to \cite{6774981}, the WCS algorithm takes into account the impact of network interference on user association {leveraging the CSI of channels between users and BSs} in a downlink scenario \cite{8677293}.
To improve the system energy efficiency combined with the network throughput, Zhang et al. deploy a novel energy-efficient mmWave based ultra-dense network optimization framework in a downlink scenario \cite{7961156}.

Since the user association in mmWave vehicular networks has to be invoked constantly as channels dramatically change, offline algorithms are not suitable or suffer from significant time complexity.
To adapt to dynamic wireless networks with affordable computation complexity, learning-based methods have been utilized for user association.
{
	A contextual MAB approach has been adopted for the optimal beam selection at mmWave BSs \cite{8472783,li2020smart}. 
	Nonetheless, given an expansive context space within mmWave vehicular networks, it is costly for these algorithms to learn adequately.
	The well-known LinUCB algorithm can learn efficiently given a large context space \cite{slivkins2019introduction}.
	Nonetheless, the LinUCB algorithm assumes that rewards satisfy Lipschitz-continuity and exhibit a linear relationship with the context, which is too stringent for mmWave vehicular networks due to the fast-fading channels and Doppler effect.}
Formulating the beam selection problem in the vehicular scenario as a combinatorial MAB problem, Nasim et al. propose a sequential combinatorial TS scheme to choose the optimal beams \cite{9286435}.
Users determine their associations considering only the interference introduced by the previous users which have already determined their associations.
Nevertheless, the Doppler effect introduced by the fast movement of vehicles is not taken into account, and only a single BS is deployed within the network.
In \cite{9789983}, Alizadeh et al. propose a central load balancer algorithm to associate users and BSs while ensuring that each BS serves a maximum number of users.
However, the algorithm is only suitable for a slow-fading channel.


Different from the above works, we propose a low-complexity online algorithm, i.e., SD-CC-UCB, which learns reward distribution without requiring any CSI.
The SD-CC-UCB algorithm can estimate the transmission rate given the location and velocity of a vehicle, which eliminates significant delays caused by frequent channel estimations.

\section{System Model}
Focusing on the user association in the mmWave vehicular network, we first provide an overview of the mobility model of vehicles, channel characteristics, and models.
Based on the network model, we then formulate the optimization problem and introduce the overall algorithm design involving the cooperation between vehicles and BSs.
\subsection{Mobility Model}
Vehicles move along the streets and update their locations during each period $t = 1,\dots,T$.
Here $T \in \mathbb{N}$ is a finite time horizon.
Let $\mathbb{U}(t) = \{1,\dots,i,\dots,|\mathbb{U}(t)|\}$, $v_i(t)$ and $l_i(t)$ denote the set of vehicles, the velocity, and location of vehicle $i$ at $t$, respectively.
The velocity $v_i(t)$ follows a uniform distribution on $[v_{min}, v_{max}]$.
The location of vehicle $i$ can be represented as $l_i(t) = (x_i(t), y_i(t))$.
We define $\mathbb{L}$ as the set of all possible locations and $\mathbb{V}$ as the set of velocities of vehicles.
The arrival of vehicles at each period follows a Poisson distribution with density $\lambda$.

\subsection{Channel Model}
Let $\mathbb{B} = \{1,\dots,j\dots,N_{BS}\}$ indicate the set of BSs.
According to the 3GPP TR 38.901 standard \cite{3gpp2018study}, each vehicle is assumed to be equipped with a rectangular antenna array.
Each BS is deployed with a massive antenna array, enabling it to serve multiple vehicles simultaneously.

In the vehicular communication network, the mobility of vehicles introduces high Doppler spreads, leading to a short coherence time \cite{zochmann2018measured}.
To portray the Doppler effect, the Clustered Delay Line (CDL) channel model with ray tracing \cite{3gpp2018study} is utilized in this paper.
{It generates wireless channels based on the geometrical characteristics of the channel and the estimated maximum Doppler spread.}
The ray tracing technique can describe the geometrical characteristics of the channel, such as the number of rays, the angle of departure and arrival, delays, and attenuation.
Based on the vehicle's velocity, the departure angles, and the arrival angles of rays, the maximum Doppler spread is estimated.
The above parameters collectively establish the CDL channel model.

In each period $t$, one vehicle can only communicate with one BS.
Let $\beta_i(t)$ indicate the index of the serving BS for vehicle $i$.
The association between BSs and vehicles can be defined as
\begin{equation}
	\boldsymbol{\beta}(t) \triangleq [\beta_1(t),\dots,\beta_i(t),\dots,\beta_{|\mathbb{U}(t)|}(t)].
\end{equation}

When BS $j$ serves vehicle $i$, the channel gain between vehicle $i$ and BS $j$ is
\begin{equation}
	\label{h_ij_0}
	h_{i,j}(t) = {(\mathbf{w}^r_{i,j})} ^*\mathbf{H}_{i,j}(t)\mathbf{w}^t_{i,j}.
\end{equation}
Here, $\mathbf{H}_{i,j}(t)$ is the channel matrix between vehicle $i$ and BS $j$.
$\mathbf{w}^r_{i,j}$ and $\mathbf{w}^t_{i,j}$ are the receive and transmit beamforming weights calculated by singular value decomposition.

Considering another vehicle $k$ transmits to BS $l$ over an uplink.
The transmission from vehicle $i$ to BS $j$ might be interfered with the concurrent transmission from vehicle $k$ to BS $l$. 
At period $t$, the perceived interference to BS $j$ from vehicle $k$ while serving vehicle $i$ is
\begin{equation}
	\label{h_ij}
	\widetilde{h}^{k,l}_{i,j}(t) = {(\mathbf{w}^r_{i,j})} ^*\mathbf{H}_{k,j}(t)\mathbf{w}^t_{k,l}.
\end{equation}
Here $\mathbf{w}^t_{k,l}$ is the transmit beamforming weight of vehicle $k$.
Assume each vehicle transmits at power $P_v$, the interference plus noise at BS $j$ while serving vehicle $i$ is
\begin{equation}
	\label{I_ij}
	I_{i,j}(t) = \Big|\sum_{k \in (\mathbb{U}(t) \backslash i)} \sum_{l \in \mathbb{B}}  P_v \widetilde{h}^{k,l}_{i,j}(t) \mathcal{I}_{k,l}(t) \Big| ^ 2 + N_oW,
\end{equation}
where $N_o$ is the thermal noise power density and $W$ is the bandwidth.
$\mathcal{I}_{k,l}(t)$ is the indicator function.
If vehicle $k$ communicates to BS $l$, $\mathcal{I}_{k,l}(t) = 1$.
Otherwise, $\mathcal{I}_{k,l}(t) = 0$.

\subsection{Optimization Problem and Algorithm Architecture}
\label{Algorithm Architecture}
Given a vehicle $i$, according to (\ref{h_ij}) and (\ref{I_ij}), the instantaneous transmission rate of vehicle $i$ communicating to BS $j$ is
\begin{equation}
	\label{bar R_ij}
	\bar R_{i,j}(t) = W\log_2(1 + \frac{P_v|h_{i,j}(t)|^2}{I_{i,j}(t)}).
\end{equation}
As the interference $I_{i,j}(t)$ and channel gain $h_{i,j}$ vary with time, a vehicle has to switch to another serving BS to sustain the transmission frequently.
During the handover, the vehicle experiences a certain decrease in the transmission rate.
{The handover cost $\zeta$ is defined as the ratio of the time spent for handover to the duration of the data transmission phase within a period.}
According to (\ref{bar R_ij}), the instantaneous transmission rate considering the handover cost is
\begin{equation}
	\label{R_ij}
	R_{i,j}(t) = (1 - \zeta\mathbb{I}_{\beta_i(t - 1) \neq j})W\log_2(1 + \frac{P_v|h_{i,j}(t)|^2}{I_{i,j}(t)}),
\end{equation}
where the indicator function $\mathbb{I}_{\beta_i(t - 1) \neq j}$ is 1 if a handover happens at period $t$, otherwise $\mathbb{I}_{\beta_i(t - 1) \neq j}$ is 0.
Then, the total instantaneous transmission rate of all vehicles is
\begin{equation}
	\label{total_transmission_rate}
	\begin{aligned}
		r(\boldsymbol{\beta}(t))& = \sum_{i \in \mathbb{U}(t)} R_{i,\beta_i(t)}(t)= \sum_{i \in \mathbb{U}(t)} (1 - \zeta\mathbb{I}_{\beta_i(t - 1) \neq \beta_i(t)})\\&\times W\log_2(1 + \frac{P_v|h_{i,\beta_i(t)}(t)|^2}{I_{i,\beta_i(t)}(t)}).
	\end{aligned}
\end{equation}

In our research, the ultimate objective of the user association is to identify an association vector which maximizes the utility of the network.
According to (\ref{total_transmission_rate}), the optimization problem at period $t$ is formulated as 
\begin{equation}
	\label{formulate_semi_CC_UCB}
	\begin{aligned}
		&\max_{\boldsymbol{\beta}} \quad r(\boldsymbol{\beta}(t))\\
		&\begin{array}{r@{\quad}r@{}l@{\quad}l}
			s.t. &\sum_{j=1}^{N_{BS}}\mathcal{I}_{i,j}(t)& = 1, \forall i \in \mathbb{U}(t).\\
		\end{array}
	\end{aligned}
\end{equation}
Recall that $\mathcal{I}_{i,j}(t) = 1$ if and only if vehicle $i$ selects BS $j$ as the serving BS.
The constraint in (\ref{formulate_semi_CC_UCB}) indicates that a BS can serve multiple vehicles simultaneously but a vehicle can only connect to one BS.

\begin{figure*}[htpb]
	\centering\includegraphics[scale=0.73]{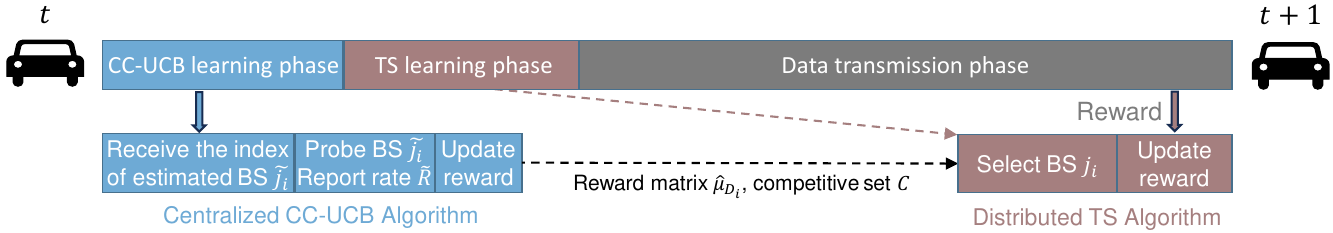}
	\caption{Architecture of SD-CC-UCB algorithm.}
	\label{structure_for_vehicle}
\end{figure*}

The optimization problem formulated in (\ref{formulate_semi_CC_UCB}) is an integer non-convex programming {which can be formulated as the Generalized Proportional Fairness (GPF1) problem.
	The GPF1 problem can be reduced from the well-known NP-hard 3-dimensional matching problem \cite{ramjee2006generalized}, so the problem in (\ref{formulate_semi_CC_UCB}) is also NP-hard}.
The SD-CC-UCB algorithm optimizes this problem relying solely on learning transmission rate, without the need of the CSI $h_{i,j}(t)$ and $\widetilde{h}^{k,l}_{i,j}(t)$ defined in (\ref{h_ij_0}) and (\ref{h_ij}), respectively.
In consideration of this, we solve this optimization problem in two steps, which correspond to the learning phases of the CC-UCB algorithm and the TS algorithm, as shown in Fig.~\ref{structure_for_vehicle}.
First, the CC-UCB algorithm determines the channel to be estimated for each vehicle.
Then, the vehicle transmits a pilot signal to the BS to be estimated, say $\widetilde{j_i}$ and measures the transmission rate $\widetilde{R}$.
The CC-UCB algorithm then updates the rewards of the estimated BSs with the transmission rate $\widetilde{R}$.
In the succeeding TS phase, the vehicle with context $D_i$ obtains the reward matrix $\boldsymbol{\hat{\mu}}_{D_i}$ and competitive arm set $\mathbf{C}$ which is the set of BSs worthy of exploring according to the CC-UCB algorithm.
Then, the vehicle makes a decision on which BS, denoted as $j_i$ within the competitive arm set $\mathbf{C}$, to communicate with during the data transmission phase.
After the transmission phase finishes, the vehicle measures the instantaneous transmission rate and updates the reward of the TS algorithm accordingly.

Now we formulate the optimization problem in the first step, namely CC-UCB learning phase.
{While vehicle $i$ transmits a pilot to the estimated BS $j$ at period $t$, it experiences interference from surrounding vehicles.
	Let the interference, denoted as ${\eta}_{i,j}(\lambda,t)$, follow a stochastic distribution ${\eta}(\lambda)$.
	The interference is defined with respect to the arrival rate of vehicles $\lambda$ because the density of vehicles significantly influences the strength of the interference.
	For a short time horizon $T$, the arrival rate of vehicles $\lambda$ is constant, and the density of vehicles is stationary within $T$.
	Therefore, the probability distribution of the interference ${\eta}(\lambda)$ remains stationary within $T$, and ${\eta}_{i,j}(\lambda,t)$ is quasi-stationary.}
{For the channel} from vehicle $i$ to the BS $j$, the transmission rate during estimation is
\begin{equation}
	\label{learning_R_ij}
	\widetilde{R}_{i,j}(t) = W\log_2(1 + \frac{P_v|h_{i,j}(t)|^2}{N_oW + {\eta_{i,j}(\lambda,t)}}).
\end{equation}
{The interference ${\eta}_{i,j}(\lambda,t)$ is determined by the association vector and current channel conditions.
	Formally, 
	\begin{equation}
		\nonumber
		{\eta}_{i,j}(\lambda,t) = \Big|\sum_{k \in (\mathbb{U}(t - 1) \backslash i)} \sum_{l \in \mathbb{B}}  P_v \widetilde{h}^{k,l}_{i,j}(t) \mathcal{I}_{k,l}(t - 1) \Big| ^ 2.
\end{equation}}
Let $\widetilde{\boldsymbol{\beta}}(t) \triangleq [\widetilde\beta_1(t),\dots,\widetilde\beta_i(t),\dots,\widetilde\beta_{|\mathbb{U}(t)|}(t)]$ indicate the learning set of CC-UCB, which is the set of the BSs to be estimated at period $t$.
The total transmission rate is 
\begin{equation}
	\label{widetilde_learning_r}
	\begin{aligned}
		\widetilde r(\boldsymbol{\widetilde\beta}(t)) &= \sum_{i \in \mathbb{U}(t)} \widetilde R_{i,\widetilde \beta_i(t)}(t)
	\end{aligned}
\end{equation}
According to (\ref{widetilde_learning_r}), the optimization problem in the CC-UCB learning phase can be formulated as
\begin{equation}
	\label{formulate_CC_UCB}
	\begin{aligned}
		&\max_{\widetilde{\boldsymbol{\beta}}} \quad \widetilde r(\widetilde{\boldsymbol{\beta}}(t)) = \sum_{i \in \mathbb{U}(t)}W\log_2(1 + \frac{P_v|h_{i,\widetilde \beta_i(t)}(t)|^2}{N_oW + {{\eta}_{i,\widetilde\beta_i(t)}(\lambda,t)}})\\
		&\begin{array}{r@{\quad}r@{}l@{\quad}l}
			s.t. &\sum_{j=1}^{N_{BS}}\mathcal{I}_{i,j}& = 1, \forall i \in \mathbb{U}(t).\\
		\end{array}
	\end{aligned}
\end{equation}

The learned transmission rate in the CC-UCB phase is not accurate because interference and handover cost are ignored.
{Also, $\eta_{i,j}(\lambda,t)$ introduces a bias to the reward in (\ref{learning_R_ij}).}
Based on the learned transmission rate and competitive BSs generated by CC-UCB,
the TS algorithm is applied to improve accuracy by rectifying the discrepancy between the estimated transmission rate and instantaneous rate in a distributed manner.
Finally, the optimization problem in (\ref{formulate_semi_CC_UCB}) is solved in a semi-distributed online algorithm.

\section{Contextual Correlated Upper Confidence Bound Algorithm}
\label{section CC-UCB}
User association relies on up-to-date channel information.
However, the channel coherence time of mmWave vehicular channels is very short \cite{7390852}, so frequent channel estimations introduce notable signaling overhead and delays.
Furthermore, the calculation of CSI is time-consuming in large scale mmWave vehicular networks \cite{9732214}.

In this section, we propose a CC-UCB algorithm based on the contextual MAB framework and the C-BANDIT {algorithm} \cite{9434422} to capture the channel conditions of the fast-fading mmWave vehicular channels.
{
	Noticing the reward distributions for an arm across different contexts are correlated, CC-UCB can gain the knowledge of reward distributions across multiple contexts by leveraging previously sampled rewards to recognize the BSs worth sampling.
	Moreover, avoiding CSI measurement and calculation, CC-UCB allows a vehicle to predict the transmission rate to each BS, significantly relieved from burdensome channel estimations.
}

\subsection{Contextual Multi-Armed Bandits}
\label{Contextual Multi-Armed Bandits}
The objective in (\ref{formulate_CC_UCB}) is non-convex, and the variables are restricted to be an integer.
Moreover, the objective is a time-varying function considering the mobility of vehicles.
Addressing user association decision problems in a highly dynamic vehicular network is extremely difficult and hardly feasible since the one-shot solution in the current period $t$ may not work well in succeeding periods as the network evolves.
Thus, we resort to an online learning algorithm to address this problem and formulate it as a contextual MAB problem.
In the contextual MAB framework, each BS is treated as an arm, and we consider a central controller as the agent.
At each period, the vehicle transmits a pilot to the estimated BS decided by the agent, obtains the transmission rate as a reward, and reports the feedback to the agent.
Then, the agent updates rewards accordingly.

{We define a context in the contextual MAB problem as the location and velocity of a vehicle.}
Given two nearby vehicles, they share similar LOS and NLOS links to the same BS due to the sparsity of the mmWave channels.
Further, if the two vehicles move at a comparable speed, they experience similar Doppler spread and the channel conditions from these two vehicles to the same BS are comparable.
{Therefore, at each context, the reward distributions are stationary because the vehicle experiences similar LOS, NLOS links and Doppler spread.}
The context space is the set of all contexts with regard to the location space $\mathbb{L}$ and velocity space $\mathbb{V}$.
The BS with the highest transmission rate to the vehicle in each context can be learned online through the contextual UCB (CUCB) algorithm.
After the agent has conducted sufficient explorations, the BS with the highest transmission rate, i.e., the optimal arm, can be predicted.
Once a vehicle enters the network, the optimal arm can be foreseen given its location and velocity without extra explorations.

Consider a contextual MAB setting with $N_{BS}$ arms.
Given vehicle $i$, at any period, the agent observes the corresponding context $D_i$ defined by the location and velocity of vehicle $i$.
Let $\hat{\mu}_{D_i,j}$ indicate the empirical reward of arm $j$ on context $D_i$
\begin{equation}
	\hat{\mu}_{D_i,j} = \sum_{n = 1}^{n_{D_i,j}}\frac{\widetilde R(D_i,j,n)}{n_{D_i,j}},
\end{equation}
where $n_{D_i,j}$ is the number of times that BS $j$ is selected given the context $D_i$ and $\widetilde R(D_i,j,n)$ is the obtained reward at the $n$th instance.
Then, the arm with the greatest UCB index $I_{D_i,j}^{UCB}$ is selected by vehicle $i$, where the UCB index is calculated by \cite{auer2002finite}
\begin{equation}
	\label{UCB_index}
	I_{D_i,j}^{UCB} = \hat{\mu}_{D_i,j} + \alpha\sqrt{\frac{2\log (n_{D_i})}{n_{D_i,j}}}.
\end{equation}
Here, $n_{D_i}$ is the number of times that context $D_i$ has been observed.
Formally, $n_{D_i} = \sum_{j \in \mathbb{B}}n_{D_i,j}$.
Parameter $\alpha$ determines whether the algorithm focuses more on exploration or exploitation.
The UCB index is high if arm $j$ has a great empirical reward or if it has been selected fewer compared to other arms.
After pulling arm $j$, the agent receives a reward and updates the empirical reward on context $D_i$ for arm $j$.
After several explorations and reward updates, the exploitation strategy employed by the CUCB algorithm will gradually converge towards the BS with the highest transmission rate.

\subsection{Contextual Correlated UCB Algorithm}
Although the user association decision problem can be modeled as a contextual MAB problem and solved with the CUCB algorithm, the location dimension of the contexts tends to be substantial given the extensive coverage area of the network. 
Furthermore, the prominent Doppler effect in the vehicle mmWave network requires fine-grained velocity resolution.
The large scale contexts demand extensive explorations to ensure that each context has been sufficiently explored.
Consequently, the expected reward of the CUCB algorithm deteriorates for a tremendous context space.

Inspired by \cite{9434422,10.1145/3492866.3549727}, we improve the {CUCB algorithm} by harnessing the correlation of reward distributions on different contexts.
First, we define contextual pseudo-reward and empirical contextual pseudo-reward.
\begin{definition}[\emph{Contextual Pseudo-reward}]
	If an agent pulls arm $j$ on context $D_1$ and observes reward $r$, then the contextual pseudo-reward of arm $j$ on context $D_2$ with regard to arm $j$ of context $D_1$, indicated by $s_j(D_1,D_2,r)$, is an upper bound on the conditional expected reward of arm $j$ on context $D_2$, that is,
	\begin{equation}
		\mathbb{E}[R^{ct}_{D_2,j}|R^{ct}_{D_1,j} = r] \le s_j(D_1,D_2,r).
	\end{equation}
	Here $R^{ct}_{D_2,j}$ and $R^{ct}_{D_1,j}$ indicates the expected reward of pulling arm $j$ on context $D_2$ and $D_1$, respectively.
\end{definition}
\begin{definition}[\emph{Empirical Contextual Pseudo-reward}]
	After $t$ rounds, let $n_j(D_1,D_2)$ denote the times that the reward of arm $j$ on context $D_1$ updates the contextual pseudo-reward of context $D_2$.
	The empirical contextual pseudo-reward, $\hat \phi_j(D_1,D_2)$, for each arm $j$ on context $D_2$ relative to context $D_1$ is
	\begin{equation}
		\label{update_empirical_cpr}
		\hat \phi_j(D_1,D_2) = \frac{\sum_{\tau = 1}^t\sum_{i = 1}^{|\mathbb{U}(t)|}s_j(D_1,D_2,\widetilde{R}(i,\tau))\mathcal{I}_j(D_1,D_2,i,\tau)}{n_j(D_1,D_2)}.
	\end{equation}
	If the reward of vehicle $i$ on context $D_1$ pulling arm $j$ updates the contextual pseudo-reward of the arm $j$ on context $D_2$ at period $\tau$, $\mathcal{I}_j(D_1,D_2,i,\tau) = 1$. Otherwise, $\mathcal{I}_j(D_1,D_2,i,\tau)$ is 0.
	$\widetilde{R}(i,\tau)$ indicates the reward of vehicle $i$ at period $\tau$.
\end{definition}
Empirical contextual pseudo-reward is the expectation of contextual pseudo-reward over $T$ periods.
Let $\boldsymbol{\kappa}$ denote the set of the contexts which have been used to update the contextual pseudo-reward for arm $j$ on context $D_i$ at least once.
For vehicle $i$ with context $D_i$ pulling arm $j$, the lowest empirical contextual pseudo-reward is
\begin{equation}
	\label{lowest empirical contextual pseudo-reward}
	\mathbb{\hat{\phi}}_{D_i,j} = \min_{D \in \boldsymbol{\kappa}}\hat \phi_j(D,D_i).
\end{equation}
Here, $D \in \boldsymbol{\kappa}$ is a context.
Based on the empirical contextual pseudo-reward and empirical reward, the set of arms is classified into non-competitive arms and competitive arms, which are defined in the following definition.
Then, the algorithm only explores competitive arms, knocking out unnecessary explorations on non-competitive arms.

\begin{definition}[\emph{Non-competitive Arm and Competitive Arm}]
	\label{com-noncom-definition}
	Consider a vehicle $i$ with context $D_i$ at period $t$.
	Let $\hat{\mu}_{{D_i}, j^e}$ indicate the greatest empirical reward on context $D_i$,
	and $j^e$ is the arm with the greatest empirical reward.
	If $\mathbb{\hat{\phi}}_{D_i,j} \ge \hat{\mu}_{{D_i}, j^e}$ or $\boldsymbol{\kappa} = \varnothing$, arm $j$ is a competitive arm, otherwise arm $j$ is a non-competitive arm.
\end{definition}
Pulling arm $j$ will not yield any improvement if its empirical contextual pseudo-reward is lower than the greatest empirical reward obtained so far.
So arm $j$ could be ruled out as a non-competitive arm to save explorations without sacrificing the reward from exploitation. 
If $\boldsymbol{\kappa}$ is an empty set, the empirical contextual pseudo-reward is unknown, so the arm needs to be explored and is considered competitive.

\begin{figure}[h] 
	\centering\includegraphics[scale=0.4]{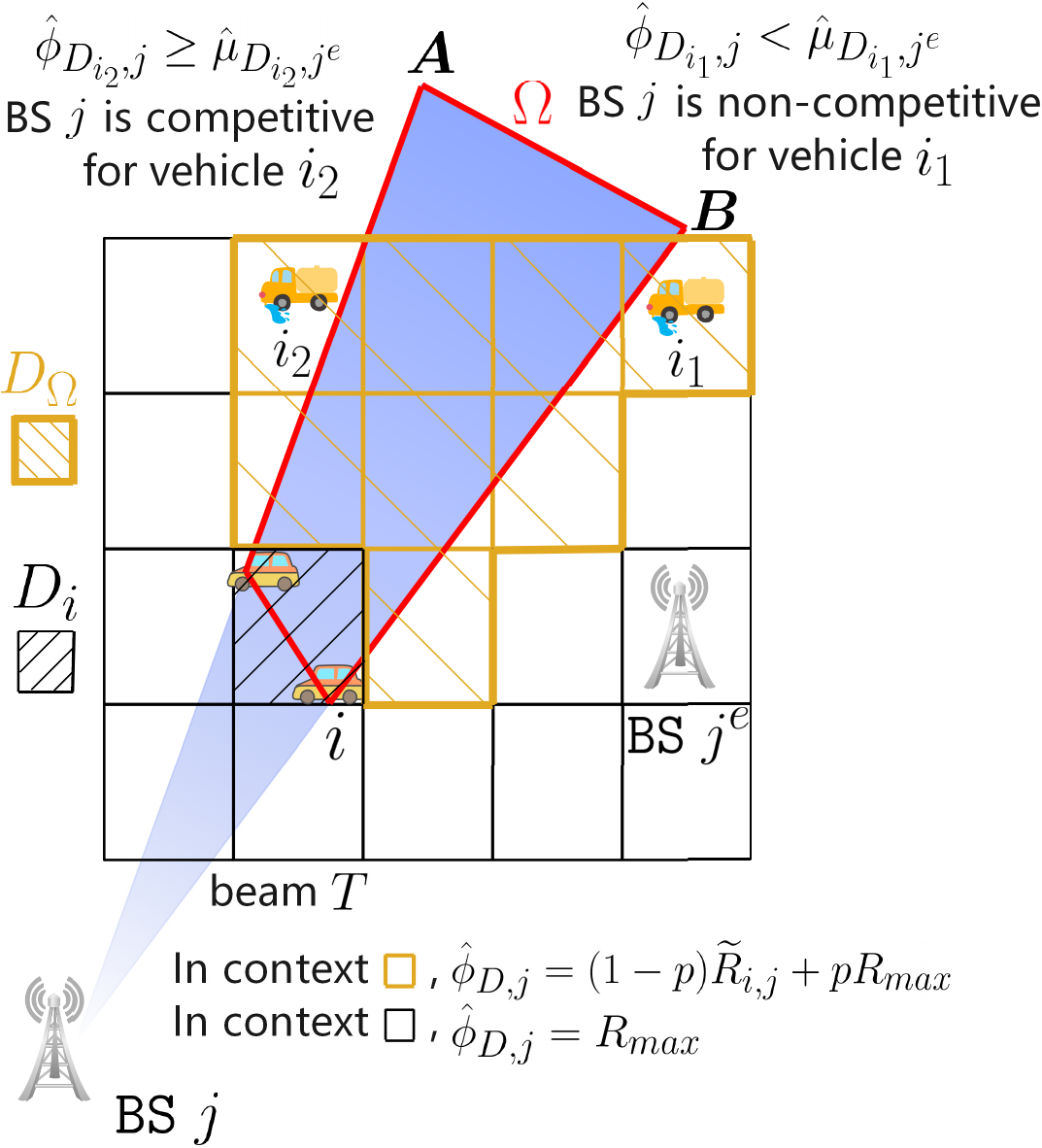}
	\caption{
		The grids represent the location dimension of contexts.
		$p$ and $R_{max}$ are the infraction probability and the highest achievable transmission rate for a vehicle, respectively.
		$\mathbb{\hat{\phi}}_{D,j}$ indicates the lowest empirical contextual pseudo-reward of arm $j$ on context $D$.
	}
	\label{Fig_CLUB}
\end{figure}

Now we investigate the contextual pseudo-reward $s_j(D_1,D_2,r)$.
Consider the following scenario: 
A vehicle $i$ on context $D_i$ communicates to BS $j$ and receives a reward $\widetilde R_{i,j}$ at period $t$ as shown in Fig.~\ref{Fig_CLUB}.
During the brief traveling time from period $t$ to $t + 1$, vehicle $i$ travels from location $l_i(t)$ to $l_i(t + 1)$ with velocity $v_i(t)$.
Consider another vehicle $i_1$ with comparable velocity is located within  $\mathbf{\Omega}(l_i(t),l_j,l_i(t + 1))$.
Here, the region $\mathbf{\Omega}(l_i(t),l_j,l_i(t + 1))$ indicates the quadrilateral with red borders which is extended from beam $T$.

During the movement, there are two situations in arm selection for two vehicles covered by the same beam $T$ of a BS.
In the first case, vehicle $i$ transmits to BS $j$ over a LOS link.
Due to the large-scale fading, vehicle $i_1$ is likely to encounter a transmission rate lower than $\widetilde R_{i,j}$ as it is farther away from BS $j$ compared to vehicle $i$.
In the second case, the LOS propagation of vehicle $i$ to BS $j$ is occluded.
Because vehicle $i_1$ is covered by the same beam $T$ of BS $j$ as vehicle $i$, the LOS link of vehicle $i_1$ is also blocked.
Compared to vehicle $i$, the signal from vehicle $i_1$ in the outer sector of beam $T$ experiences heavier path loss due to the large-scale fading.
As a result, vehicle $i_1$ encounters a poorer channel gain and experiences a lower transmission rate compared to vehicle $i$.
In conclusion, the transmission rate of a vehicle with a certain context is bounded by $\widetilde R_{i,j}$ with high probability.
This conclusion can be applied to determine the contextual pseudo-reward of arm $j$ on the contexts with the location dimensions inside the beam $T$.

Given velocity $v_i(t)$, let $\mathbb{D}_{\mathbf{\Omega}}$ indicate the set of contexts where a vehicle is located in the same beam $T$ as vehicle $i$ but possibly encounters a transmission rate below $\widetilde R_{i,j}$, as the contexts in orange shade illustrated in Fig.~\ref{Fig_CLUB}. Formally,
\begin{equation}
	\nonumber\begin{aligned}
		\mathbb{D}_{\mathbf{\Omega}} \triangleq &\{D = (\mathbf{L},\mathbf{V}) \in \mathbb{D} \backslash D_i:\\
		&\mathbf{L} \cap \mathbf{\Omega}(l_i(t),l_j,l_i(t + 1)) \neq \varnothing, v_i(t) \in \mathbf{V}\}.
	\end{aligned}
\end{equation}
Here, $\mathbf{L}$ and $\mathbf{V}$ are the location and velocity dimensions of context $D$, respectively.
However, attributed to small-scale fading {and the biased reward subject to interference during the estimation according to (\ref{learning_R_ij})}, vehicle $i_1$ probably achieves a transmission rate higher than $\widetilde R_{i,j}$ occasionally.
Let $p$ indicate the infraction probability that vehicle $i_1$ experiences a transmission rate over $\widetilde R_{i,j}$, the contextual pseudo-reward of arm $j$ on context $D$ with regard to arm $j$ on context $D_i$ is derived as
\begin{equation}
	\label{pseudo_reward_func}
	\begin{aligned}
		s_j(D_i,D,\widetilde R_{i,j}) = \begin{cases}
			(1 - p) \widetilde R_{i,j} + p R_{max},&D \in \mathbb{D}_{\mathbf{\Omega}},\\
			R_{max},& \text{otherwise}.
		\end{cases}
	\end{aligned}
\end{equation}
$R_{max}$ represents the highest achievable transmission rate for a vehicle across the network.
If vehicle $i_1$ is located outside $\mathbf{\Omega}(l_i(t),l_j,l_i(t + 1))$, as the contexts in black shade shown in Fig.~\ref{Fig_CLUB}, there is no straightforward correlation between vehicle $i_1$ and vehicle $i$.
To derive a tight upper bound of the reward, the upper bound of the transmission rate of the vehicle outside $\mathbf{\Omega}(l_i(t),l_j,l_i(t + 1))$ is set to the maximum achievable reward $R_{max}$.

\begin{algorithm}[htbp]
	
	\caption{{CC-UCB Estimating Phase}}
	\label{CC_UCB}
	
	\KwIn{$\boldsymbol{\hat \mu}$, $\boldsymbol{\hat \phi}$, $\boldsymbol{\hat{\phi}_{inf}}$, $t$, $i$, $d_i(t)$} 
		
		
			
			
			Find context $D_i$ that satisfies $d_i(t) \in D_i \in \mathbb{D}$
			
			Calculate $n_{D_i},\mathbb{S}_{D_i}, j^e$
			
			Initialize empirically competitive set $E_i(t) = \{j^e\}$
			
			\For{$j \in \mathbb{B} \backslash j^e$}{
				
				Find set $\boldsymbol{\kappa}$ and calculate $\mathbb{\hat{\phi}}_{D_i,j}$
				
				\If{$\mathbb{\hat{\phi}}_{D_i,j} \ge \hat{\mu}_{{D_i}, j^e}$ \textbf{or} $\boldsymbol{\kappa} = \varnothing$}{
					$E_i(t) = E_i(t) \cup \{j\}$
				}
			}
			$\widetilde{j}_i = \arg\max_{j\in E_i(t)}I_{D_i,j}^{UCB}$
			
			\KwOut{Estimated BS $\widetilde{j_i}$, reward matrix $\boldsymbol{\hat{\mu}}_{D_i}$, competitive set $E_i$ for vehicle $i$}
		
	\end{algorithm}
	
	\begin{algorithm}[htbp]
		\caption{{CC-UCB Updating Phase}}
		\label{CC_UCB_update} 
		\KwIn{Reward $\widetilde{R}_{i,\widetilde{j}_i}(t)$ and the corresponding context $D_i$ for each vehicle $i$ at period $t$}
		\For{each vehicle $i$}{
			Update $n_{D_i,\widetilde{j}_i} = n_{D_i,\widetilde{j}_i} + 1$
			
			Update empirical reward $\hat{\mu}_{D_i,\widetilde{j}_i} = \frac{\hat{\mu}_{D_i,\widetilde{j}_i}(n_{D_i,\widetilde{j}_i} - 1) + \widetilde{R}_{i,\widetilde{j}_i}(t)}{n_{D_i,\widetilde{j}_i}}$
			
			Update the UCB Index according to (\ref{UCB_index})
			
			Update empirical contextual pseudo-rewards using CLUB rule (Algorithm \ref{update_Pseudo})
		}
		\KwOut{$\boldsymbol{\hat \mu}$, $\boldsymbol{\hat \phi}$, $\boldsymbol{\hat{\phi}_{inf}}$} 
\end{algorithm}

Based on the contextual pseudo-reward derived in (\ref{pseudo_reward_func}), we propose a correlated location-based upper bound (CLUB) updating rule to update empirical contextual pseudo-reward according to (\ref{update_empirical_cpr}).
CLUB finds out $\mathbb{D}_{\mathbf{\Omega}}$ and only updates the empirical contextual pseudo-reward with context $D\in\mathbb{D}_{\mathbf{\Omega}}$, since other contexts are not evidently correlated with $D_i$.

{The proposed CC-UCB algorithm is composed of estimating phase and updating phase (Algorithm \ref{CC_UCB} and Algorithm \ref{CC_UCB_update}).}
First, CC-UCB uniformly partitions the three dimensional context space $(\mathbb{V},\mathbb{L}) \in \mathbb{R}^3$, i.e.,$\mathbb{V}$ is divided into $N_v$ intervals and $\mathbb{L}$ is divided into $N_x \times N_y$ rectangles. 
$\mathbb{D}$, $\boldsymbol{\hat \mu}$ and $\boldsymbol{\hat \phi}$ are defined as the resulting partitioned context, the empirical reward matrix, and the empirical contextual pseudo-reward matrix.
{At period $t$, each vehicle $i$ observes $d_i(t) = (v_i(t), x_i(t), y_i(t))$ and reports it to the associated BS from the last period.
	The BS performs Algorithm \ref{CC_UCB} and informs vehicle $i$ its estimated BS, reward matrix and competitive set.}
In Algorithm \ref{CC_UCB}, the arm with the highest empirical reward $j^e$ is utilized to distinguish empirically competitive and non-competitive arms, so it needs to be adequately sampled.
To guarantee that arm $j^e$ has been sampled sufficiently, we define $\mathbb{S}_{D_i}$ as the set of arms with at least $\lfloor\frac{n_{D_i}}{N_{BS}}\rfloor$ samples and then identify the optimal arm $j^e$ within $\mathbb{S}_{D_i}$,
\begin{equation}
	j^e = \arg\max_{j \in \mathbb{S}_{D_i}} \hat{\mu}_{D_i,j}.
\end{equation}
In lines 4-9, CC-UCB classifies the empirically competitive and non-competitive arms.
{Note that calculating $\hat{\phi}_{D_i,j}$ for all BSs according to (\ref{lowest empirical contextual pseudo-reward}) yields a computational complexity of $\mathcal{O}(N_{BS}|\mathbb{D}|)$.
	To reduce the complexity, we define a matrix $\boldsymbol{\hat{\phi}_{inf}}$ to represent the lowest empirical contextual pseudo-rewards of each context and $\boldsymbol{\hat{\phi}_{inf}}$ is updated in the CLUB rule (Algorithm \ref{update_Pseudo}).
	$\hat{\phi}_{D_i,j}$ is updated as $\boldsymbol{\hat{\phi}_{inf}}(D_i,j)$ only if a new sample has updated the empirical contextual reward of context $D_i$, instead of computing $\hat{\phi}_{D_i,j}$ according to (\ref{lowest empirical contextual pseudo-reward}) at each period.
	Accordingly, the BS can query the matrix $\boldsymbol{\hat{\phi}_{inf}}$ to obtain $\hat{\phi}_{D_i,j}$, significantly decreasing the latency involved in determining the estimated BS.
	Then, the BS with the highest UCB index in the competitive set is identified as the estimated BS.}
By solely exploring competitive arms, the CC-UCB algorithm effectively lessens fruitless explorations to reduce regret.

{After transmitting a pilot to the estimated BS, each vehicle receives a reward and sends it back to the BS.
	Then, each BS broadcasts the samples received from its associated vehicles to other BSs within the network.
	Utilizing the samples from the associated vehicles and other BSs, each BS implements Algorithm \ref{CC_UCB_update} to update the empirical rewards and UCB indices.}
{Additionally, both the empirical contextual pseudo-rewards and lowest empirical contextual pseudo-rewards are updated with the CLUB rule (Algorithm \ref{update_Pseudo}).}
\begin{algorithm}[htbp]
	\caption{CLUB Rule}
	\label{update_Pseudo}
	\KwIn{Parameters $l_i(t), l_i(t+1), v_i(t), \widetilde{R}_{i,\widetilde{j}_i}(t), D_i, \widetilde{j}_i$}
	Create the area $\mathbf{\Omega}(l_i(t),l_{\widetilde{j}_i},l_i(t + 1))$ and find $\mathbb{D}_{\mathbf{\Omega}}$
	
	\For{$D \in \mathbb{D}_{\mathbf{\Omega}}$}{
		
		Update $n_{\widetilde{j}_i}(D_i,D) = n_{\widetilde{j}_i}(D_i,D) + 1$
		
		Derive contextual pseudo-reward $s$ following (\ref{pseudo_reward_func})
		
		Update empirical contextual pseudo-reward $\hat{\phi}_{\widetilde{j}_i}(D_i,D) = \frac{(n_{\widetilde{j}_i}(D_i,D) - 1) \hat{\phi}_{\widetilde{j}_i}(D_i,D) + s}{n_{\widetilde{j}_i}(D_i,D)}$
		
		{
			\If {$\hat{\phi}_{\widetilde{j}_i}(D_i,D) < \hat{\phi}_{inf}(D,\widetilde{j}_i)$}{
				$\hat{\phi}_{inf}(D,\widetilde{j}_i) = \hat{\phi}_{\widetilde{j}_i}(D_i,D)$}
		}
	}
	\KwOut{$\boldsymbol{\hat \phi}$, $\boldsymbol{\hat{\phi}_{inf}}$}
\end{algorithm}

\subsection{Regret Analysis}
\label{CC-UCB regret Analysis}
The cumulative regret of the CC-UCB algorithm is quantified as the expected difference between the transmission rates achieved by the optimal strategy and the CC-UCB algorithm as follows,
\begin{equation}
	\label{regret_equation}
	\mathbb{E}[\widetilde{\mathcal{R}}(T)] = \mathbb{E}\Big[\sum_{t = 1}^T\sum_{i \in \mathbb{U}(t)} (\widetilde R_{i,\widetilde\beta_i^*(t)}(t) - \widetilde R_{i,\widetilde\beta_i(t)}(t))\Big].
\end{equation}
Here, $\widetilde{\boldsymbol{\beta}}^*(t) \triangleq [\widetilde\beta^*_1(t),\dots,\widetilde\beta^*_i(t),\dots,\widetilde\beta^*_{|\mathbb{U}(t)|}(t)]$ indicates the optimal association at period $t$.

{In this section, we first derive the regret for an individual context where the reward distribution is stationary and subsequently generalize the results across multiple contexts.
	Finally, we discuss the value of $|\mathbf{C}|$, demonstrating that the CC-UCB algorithm achieves moderate regret in mmWave vehicular networks.}
For a specific context $D \in \mathbb{D}$ over time horizon $T$, let $\mathcal{R_D}(T,k)$ indicate the regret of an arm $k$, the expected regret can be expressed as
\begin{equation}
	\mathbb{E}[\mathcal{R_D}(T,k)] = \mathbb{E}[n_{D,k}(T)]\Delta_{D,k},
\end{equation}
where $\Delta_{D,k} = \mu_{D,k^*} - \mu_{D,k}$ is the sub-optimality gap of arm $k$ concerning the optimal arm $k^*$, and $n_{D,k}(T)$ is the number of times that arm $k$ is pulled.
Given $N_{BS}$ available arms, the upper bound of the expected cumulative regret, denoted by $\mathbb{E}[\mathcal{R}_D(T)]$, is bounded by
\begin{equation}
	\label{R_DT}
	\begin{aligned}
		\mathbb{E}[\mathcal{R}_D(T)] & = \sum_{c \in \mathbf{C}\backslash k^*} \mathbb{E}[\mathcal{R_D}(T,c)] + \sum_{j\in \mathbb{B}\backslash\mathbf{C}} \mathbb{E}[\mathcal{R_D}(T,j)] 
		\\&\overset{\text{(a)}}{\le} \mathbb{E}[\Delta_k]\big((|\mathbf{C}| - 1) \mathcal{O}(\log TU) + \mathcal{O}(1)\big).
	\end{aligned}
\end{equation}
Here $\mathbf{C} \subseteq \mathbb{B}$ is the set of the competitive arms.
In the correlated UCB algorithm, a non-competitive arm is expected to be pulled with $\mathcal{O}(1)$ times, while a competitive arm is expected to be pulled with $\mathcal{O}(\log T)$ times \cite{9434422}.
Over $T$ periods, CC-UCB has pulled arms with $\mathcal{O}(TU)$ times, where $U$ is denoted as the number of vehicles within the network.
Inequality $(a)$ follows from this result.
Assuming contexts are observed by vehicles with equal probability, the expected sub-optimality gap $\mathbb{E}[\Delta_k]$ is bounded by
\begin{equation}
	\mathbb{E}[\Delta_k] = \sum_D \frac{\mu_{D,k^*} - \mu_{D,k}}{|\mathbb{D}|} \le \sum_D \frac{\mu_{D,k^*}}{|\mathbb{D}|} \le \mathbb{E}[\mu_{k^*}],
\end{equation}
where $\mathbb{E}[\mu_{k^*}]$ is the expected transmission rate of a vehicle when connecting to the optimal BS $k^*$ across the context space, which is a constant.
Then, the regret in (\ref{R_DT}) is bounded by
\begin{equation}
	\label{R_DT2}
	\mathbb{E}[\mathcal{R}_D(T)] \le (|\mathbf{C}| - 1) \mathcal{O}(\log TU) + \mathcal{O}(1)
\end{equation}

Let $n_D$ be the times that context $D$ is observed.
As $\sqrt{n_D} \ge 1 $, the upper bound of the regret in (\ref{regret_equation}) is
\begin{equation}
	\label{R_T}
	\begin{aligned}
		\mathbb{E}[\widetilde{\mathcal{R}}(T)]
		& = \sum_D\mathbb{E}[\mathcal{R}_D(T)] \le \mathcal{O}(1) + (|\mathbf{C}| - 1) \times \\& \mathcal{O}\big((\sqrt{n_{D_1}} + \dots + \sqrt{n_{D_{|\mathbb{D}|}}})\log TU\big) \\
		& \overset{\text{(b)}}{\le} (|\mathbf{C}| - 1) \mathcal{O}(\sqrt{|\mathbb{D}|TU}\log TU) + \mathcal{O}(1).\\ 
	\end{aligned}
\end{equation}
Inequality (b) follows from $\sum_D n_D = TU$ and $|\mathbb{D}|$ is the size of the contexts.

{
	According to (\ref{R_T}), the regret depends on the size of the competitive arms set $|\mathbf{C}|$.
	For a specific location $l$ within context $D$, the distance between a competitive BS $b$ to location $l$ is bounded by $d_{max}$, where $d_{max}$ is the maximum transmission distance between a BS and a vehicle.
	Otherwise, both the reward and contextual pseudo-reward become zero as the vehicle at $l$ cannot reach the competitive BS.
	%
	Consequently, if the distance between location $l$ and BS $b$ exceeds $d_{max}$, the contextual pseudo-reward of BS $b$ on context $D$ is lower than the highest empirical reward, and BS $b$ is identified as non-competitive.}

{Given $F_{l \rightarrow b}(r,\lambda_{BS})$ as the cumulative density function of the distance between location $l$ and BS $b$,} the expected number of BSs within the transmission range is $\pi d^2_{max}\lambda_{BS} F_{l \rightarrow b}(d_{max},\lambda_{BS})$, thus the size of the competitive arms set $|\mathbf{C}|$ can be bounded by 
\begin{equation}
	\label{C_size}
	|\mathbf{C}| \le \pi d^2_{max}\lambda_{BS} F_{l \rightarrow b}(d_{max},\lambda_{BS}).
\end{equation}

{
	According to (\ref{R_T}) and (\ref{C_size}), a short maximum transmission distance and a low density of BSs lead to small $|\mathbf{C}|$ and low regret.
	MmWave communication typically involves short transmission ranges due to significant path loss, while BSs are spread out to increase resource efficiency.
	This implies that our algorithm can achieve a low regret in mmWave vehicular networks.}

Considering a widely adopted deployment of BSs, such as Poisson point process, the cumulative density function of the distance between location $l$ and a BS $b$ is $F_{l \rightarrow b}(r,\lambda_{BS}) = 1 - \exp(-\pi\lambda_{BS} d^2_{max})$ \cite{7511676}.
According to (\ref{R_T}), the regret is bounded by
\begin{equation*}
	\begin{aligned}
		\mathbb{E}[\widetilde{\mathcal{R}}(T)] \le &\Big(\pi d^2_{max}\lambda_{BS}\big(1 - \exp(-\pi d^2_{max}\lambda_{BS} )\big) - 1\Big) \times\\& \mathcal{O}(\sqrt{|\mathbb{D}|TU}\log TU) + \mathcal{O}(1).
	\end{aligned}
\end{equation*}

\subsection{Complexity Analysis}
\label{CC-UCB complexity Analysis}
{The Algorithm \ref{CC_UCB} only takes a few scalar multiplication and addition operations.
	The computation complexity of Algorithm \ref{CC_UCB_update} is dominated by updating the pseudo-reward (Algorithm~\ref{update_Pseudo}).
	For each vehicle at a period, the computation cost of the CLUB rule is $\mathcal{O}(|\mathbb{D}|)$.
	Since the expected number of vehicles within the network is $U$, the complexity of the CC-UCB algorithm is $\mathcal{O}(|\mathbb{D}|U\big)$.}


\section{Semi-distributed Contextual Correlated Upper Confidence Bound Algorithm}
The CC-UCB algorithm can capture the channel conditions of the fast-fading mmWave vehicular channel promptly by leveraging the dependency of reward distributions on different contexts.
Nevertheless, in the dense mmWave vehicular wireless networks, the predicted transmission rate may lose accuracy without precise information about the intense time-varying interference.
Meanwhile, due to the rapid and abrupt variations in the mmWave channel, frequent handovers result in considerable signaling overhead and adversely affect the transmission rate.
Thus, based on the transmission rate and the empirical competitive set estimated with the CC-UCB algorithm, we further propose the SD-CC-UCB algorithm which is immune to interference and handover.
\subsection{SD-CC-UCB Algorithm}
\label{SD-CC-UCB Algorithm}
In contrast to the problem in (\ref{formulate_CC_UCB}), the optimization problem in (\ref{formulate_semi_CC_UCB}) is particularly complicated because of the interwoven interference between vehicles {and is NP-hard \cite{ramjee2006generalized}}.
Prior work has proposed a centralized algorithm that limits the number of users served by each BS to decrease traffic overload and interference \cite{9789983}.
However, the centralized algorithm is computationally expensive, and vehicles must wait for the index of the associated BS from the centralized controller before transmission, causing significant delays or even obsolete decisions.
Hence, we employ a low-complexity distributed online learning algorithm, i.e., the TS algorithm, for each vehicle to decide the BS to associate with, eliminating the computation and signaling delay in a centralized algorithm.
TS is a heuristic approach designed to address the exploration-exploitation trade-off in the MAB problem.
In each round, a known distribution generates a sample, and the arm with the greatest sample is pulled.
After receiving a reward, the distribution parameters of the pulled arm are updated.

The reward of the TS is defined as the discrepancy between the transmission rate estimated by CC-UCB and the instantaneous transmission rate ignoring handover cost.
{(\ref{learning_R_ij}) shows that interference impacts the reward.
	Because the distribution of interference ${\eta}(\lambda)$ remains stationary within $T$ as described in Section \ref{Algorithm Architecture}, the reward distribution is also stationary within $T$, complying with the TS algorithm's assumption of stationary reward distributions.}
For vehicle $i$, we denote $\hat\mu^{TS}_j(t)$ as the empirical reward of the BS $j$ in TS at period $t$, formally,
\begin{equation}
	\begin{aligned}
		\hat\mu^{TS}_j(t) &= \frac{1}{n_j(t)}\sum_{\tau=1}^{t}\mathbb{I}_{\beta_i(\tau) = j}\big(\hat\mu_{D(\tau),j}(\tau) - \bar{R}_{i,j}(\tau)\big) \\
		&=  \frac{1}{n_j(t)}\sum_{\tau=1}^{t}\mathbb{I}_{\beta_i(\tau) = j}(\hat\mu_{D(\tau),j}(\tau) - \frac{R_{i,j}(\tau)}{1 - \zeta\mathbb{I}_{\beta_i(\tau - 1) \neq j}}).
	\end{aligned}
\end{equation}
Recall that $\zeta$ represents the handover cost.
Let $D(\tau)$ indicate the context of the vehicle at period $\tau$.
$\hat\mu_{D(\tau),j}(\tau)$ is the transmission rate estimated by CC-UCB.
If and only if a handover occurs at period $\tau$, $\mathbb{I}_{\beta_i(\tau - 1) \neq j}$ is $1$.
Let $n_j(t) = \sum_{\tau=1}^{t}\mathbb{I}_{\beta_i(\tau) = j}$ represent the times that arm $j$ is pulled.
If vehicle $i$ communicates to BS $j$ at period $\tau$, $\mathbb{I}_{\beta_i(\tau) = j} = 1$. Otherwise, $\mathbb{I}_{\beta_i(\tau) = j}$ is $0$.
The transmission rate to BS $j$ at period $t$ is predicted by SD-CC-UCB as
\begin{equation}
	\label{S_j}
	S_{j}(t) = (\hat\mu_{D(t),j}(t) - \hat{\mu}^{TS}_j(t)) \times (1 - \zeta\mathbb{I}_{\beta_i(t - 1) \neq j}).
\end{equation}
The TS samples from the Gaussian distribution assumed with each BS within the competitive set $\mathbf{C}$.
The user then connects to BS with the greatest sample.
Let $\beta_i(t)$ indicate the BS associated with vehicle $i$ at period $t$,
\begin{equation}
	\label{find_TS_BS}
	\beta_i(t) = \arg\max_{j \in \mathbf{C}} Z_{j}, Z_{j} \sim \mathcal{N}(S_j(t),\sigma_j).
\end{equation}
The mean and variance of the Gaussian distribution of BS $j$ are $S_j(t)$ and $\sigma_j = \frac{\alpha^{TS}}{\hat{n}^{TS}_j + 1}$, respectively.
Here, $\alpha^{TS}$ and $\hat{n}^{TS}_j$ are the parameters of the TS and the count of pulling BS $j$, respectively.
The algorithm tends to explore more given a large $\alpha^{TS}$ and exploit more given a small $\alpha^{TS}$.
Only the BSs within the competitive set are sampled, as the non-competitive BSs perform poorly.
After measuring the instantaneous transmission rate during the transmission period, the vehicle updates the empirical reward of arm $j$.

\begin{algorithm}[htbp]
	\caption{{SD-CC-UCB Algorithm}}
	\label{Thompson_sampling}
	\KwIn{$\alpha^{TS}$, $t$}
		Receive  the reward matrix $\boldsymbol{\hat{\mu}}_{D(t)}$ and the up-to-date competitive arms set $\mathbf{C}$ from a BS (Algorithm~\ref{CC_UCB})
		
		\For{each BS $j$ in $\mathbf{C}$}{
			Calculate $S_{j}(t)$ and $\sigma_j$
		}
		Communicate with BS $\beta_i(t)$ derived in (\ref{find_TS_BS}) and receive reward $R$
		
		Update the counter $\hat{n}^{TS}_{\beta_i(t)} = \hat{n}^{TS}_{\beta_i(t)} + 1$
		
		Update the empirical reward $\hat{\mu}^{TS}_{\beta_i(t)} =  \frac{\hat{\mu}^{TS}_{\beta_i(t)}(\hat{n}^{TS}_{\beta_i(t)} - 1) + (\hat\mu_{D(t),\beta_i(t)} - \frac{R}{1 - \zeta\mathbb{I}_{\beta_i(t - 1) \neq \beta_i(t)}})}{\hat{n}^{TS}_{\beta_i(t)}}$
\end{algorithm}

The proposed TS algorithm is presented in Algorithm \ref{Thompson_sampling}.
Once a vehicle enters the network, the reward vector $\boldsymbol{\hat{\mu}}^{TS} = \{\hat{\mu}^{TS}_1, \dots, \hat{\mu}^{TS}_{N_{BS}}\}$ and counter vector $\boldsymbol{\hat{n}}^{TS} = \{\hat{n}^{TS}_1,\dots,\hat{n}^{TS}_{N_{BS}}\}$ are initialized.
{For each period $t$, CC-UCB (Algorithm~\ref{CC_UCB}) informs the vehicle the reward matrix $\boldsymbol{\hat{\mu}}_{D(t)} \subseteq \boldsymbol{\hat{\mu}}$ and competitive arms set $\mathbf{C}$.}
Here, $D(t)$ indicates the context of the vehicle at period $t$.
Then, the vehicle communicates with BS $\beta_i(t)$ derived in (\ref{find_TS_BS}) within the competitive set $\mathbf{C}$ for data transmission.
Upon receiving the reward, both the empirical reward and counter are updated accordingly.

\subsection{Complexity and Regret Analysis}
Computing the distribution statistics of TS and updating the reward and counter vectors only introduce scalar multiplication operations, the computation complexity of the SD-CC-UCB algorithm is dominated by updating the parameters of the CC-UCB algorithm, which is $\mathcal{O}(|\mathbb{D}|U)$ shown in Section \ref{CC-UCB complexity Analysis}.

Let $\boldsymbol{\beta}^*(t) \triangleq [\beta^*_1(t),\dots,\beta^*_i(t),\dots,\beta^*_{|\mathbb{U}(t)|}(t)]$ indicate the optimal association at period $t$.
The cumulative regret of all vehicles over $T$ periods is the expected cumulative discrepancy between the transmission rate attained by the optimal strategy and the SD-CC-UCB algorithm,
\begin{equation}
	\label{semi_CC-UCB regret_equation}
	\mathbb{E}[\mathcal{R}(T)] = \mathbb{E}\Big[\sum_{t = 1}^T\sum_{i \in \mathbb{U}(t)} (R_{i,\beta_i^*(t)}(t) - R_{i,\beta_i(t)}(t))\Big].
\end{equation}

Given a vehicle $i$, we first investigate the distribution of the reward $R_{i,\beta_i(t)}$ at period $t$.
According to (\ref{R_ij}) and (\ref{I_ij}),
\begin{equation}
	\label{regret_for_SD-CC-UCB_1}
	\begin{aligned}
		R_{i,\beta_i(t)}
		&= (1 - \zeta\mathbb{I}_{\beta_i(t - 1) \neq \beta_i(t)})W\log_2(1 + \frac{P_v|h_{i,\beta_i(t)}(t)|^2}{I_{i,\beta_i(t)}(t)})\\&
		\overset{\text{(c)}}{\approx} (1 - \zeta\mathbb{I}_{\beta_i(t - 1) \neq \beta_i(t)})W\log_2(\frac{P_v|h_{i,\beta_i(t)}(t)|^2}{I_{i,\beta_i(t)}(t)})\\
		&= (1 - \zeta\mathbb{I}_{\beta_i(t - 1) \neq \beta_i(t)})(W\log_2(\frac{P_v|h_{i,\beta_i(t)}(t)|^2}{N_oW + {\eta}_{i,\widetilde\beta_i(t)}(U,t)})\\& - W\log_2(1 + \frac{\Big|\sum\limits_{k \in (\mathbb{U}(t) \backslash i)} \sum\limits_{l \in \mathbb{B}}  P_v \widetilde{h}^{k,l}_{i,\beta_i(t)}(t) \mathcal{I}_{k,l}(t) \Big| ^ 2}{N_oW + {\eta}_{i,\widetilde\beta_i(t)}(U,t)}))\\
		&\approx (1 - \underbrace{\zeta\mathbb{I}_{\beta_i(t - 1) \neq \beta_i(t)}}_{\text{Handover cost}})
		(\underbrace{\widetilde R_{i,\beta_i(t)}(t)}_{\text{Reward in CC-UCB}} -\ \ 
		\underbrace{R^{TS}_{i,\beta_i(t)}(t)}_{\text{Reward in TS}}\ ).\\
	\end{aligned}
\end{equation}

The vehicle prefers the BS with a high estimated transmission rate $S_j$, so the SINR at the associated BS $\beta_i(t)$ is also high, i.e., $\frac{P_v|h_{i,\beta_i(t)}(t)|^2}{I_{i,\beta_i(t)}(t)} \gg 1$. Inequality (c) follows from this result.
The reward decomposes into three parts:
(1) Handover overhead.
The handover overhead is predetermined based on the previously associated BS $\beta_i(t - 1)$ with the known handover cost $\zeta$.
Therefore, this part does not introduce any regret.
(2) The reward $\widetilde R_{i,\beta_i(t)}(t)$ learned and estimated by CC-UCB, which is given by (\ref{R_T}) in Section \ref{CC-UCB regret Analysis}.
(3) The reward of TS, denoted as $R^{TS}_{i,\beta_i(t)}(t)$.
{For stationary reward distributions}, it has been demonstrated that the TS algorithm ensures an expected regret bounded by $\mathcal{O}(\sqrt{|\mathbf{C}|T\log T})$ at each agent \cite{slivkins2019introduction,cesa2006prediction}.
Substituting (\ref{R_T}) into (\ref{regret_for_SD-CC-UCB_1}), the regret in (\ref{semi_CC-UCB regret_equation}) is bounded as
\begin{equation}
	\begin{aligned}
		\mathbb{E}[\mathcal{R}(T)] \le \underbrace{(|\mathbf{C}| - 1) \mathcal{O}(\sqrt{|\mathbb{D}|TU}\log TU) + \mathcal{O}(1)}_{\text{Regret of CC-UCB}}
		\\+ \underbrace{\mathcal{O}(U\sqrt{|\mathbf{C}|T\log T})}_{\text{Regret of TS}}, 
	\end{aligned}
\end{equation}
where $|\mathbf{C}|$ is the size of the competitive arms and is bounded as $|\mathbf{C}| \le \pi d^2_{max}\lambda_{BS} F_{l \rightarrow b}(d_{max},\lambda_{BS})$.
{As discussed in Section \ref{CC-UCB regret Analysis}, $|\mathbf{C}|$ tends to be low in mmWave vehicular networks, so the SD-CC-UCB algorithm can achieve a low regret.}

\section{Numerical Results}

The simulation scenario (e.g. blockages and roads) is designed referring to a specific region in China, specifically Yuexiu District, Guangzhou. 
The coordinates of the bottom left and upper right corners of this region, i.e., (23.1280N, 113.2613E) and (23.1406N, 113.2770E), determine the simulation area.
Based on the landscape information from OpenStreetMap \cite{OpenStreetMap}, the wireless channel is emulated using the CDL channel model with ray tracing, considering the static blockages caused by the buildings.
According to the 3GPP standard \cite{3gpp2018study}, data transmissions are modulated using the widely adopted orthogonal frequency-division multiplexing (OFDM) with 135 and 65 subcarriers corresponding to the bandwidths of 100MHz and 50MHz, respectively.
BSs are uniformly deployed on the field with a height of 10 meters, according to the 3GPP TR 38.901 standard for the urban micro street canyon scenario \cite{3gpp2018study}.
The velocities of vehicles change every second, following a uniform distribution between 20 km/h and 80 km/h.
Note that the mobility of vehicles introduces the Doppler effect in the simulated channel.
The vehicle travels along streets with the shortest distance.
{We assume for the worst case that all users share the same bandwidth, which introduces intra-cell and inter-cell interference.}
Table \ref{parameter_simulation} shows the main simulation parameters.

\begin{table}[htbp]
	\caption{PARAMETERS OF PERFORMANCE EVALUATION}
	\begin{center}
		\begin{tabular}{|l|c|}
			\hline
			\hline
			\textbf{Parameter} & \textbf{Value} \\
			\hline
			Carrier Frequency & 28GHz\\
			\hline
			Transmit Power & 30dBm\\
			\hline
			Vehicle's Velocity & [20km/h, 80km/h]\\
			\hline
			Channel Bandwidth & 50MHz or 100MHz\\
			\hline
			Height of BS's Antenna & 10m \\
			\hline
			Height of Vehicle's Antenna & 2m \\
			\hline
			Infraction Probability $p$ & 0.1\\
			\hline
			Parameter for CC-UCB $\alpha$ & 1\\
			\hline
			Parameter for TS $\alpha^{TS}$& 1\\
			\hline
			Vehicle's Arrival Rate & 0.3\\
			\hline
			Number of BSs $N_{BS}$ & 9\\
			\hline
		\end{tabular}
		\label{parameter_simulation}
	\end{center}
\end{table}
Our proposed algorithms are compared to other well-known algorithms, i.e., TS, CUCB, and WCS algorithm \cite{8677293}.
TS is a traditional MAB algorithm that hinges on the assumption that the reward distribution is stationary.
CUCB can capture the underlying characteristics of the non-stationary reward distribution, but it converges slowly due to the tremendous context space.
WCS is a centralized offline algorithm and requires the instantaneous CSI between all users and BSs.

\begin{figure*}[t]
	\begin{minipage}[t]{0.33\textwidth}
		\includegraphics[width=5.7cm]{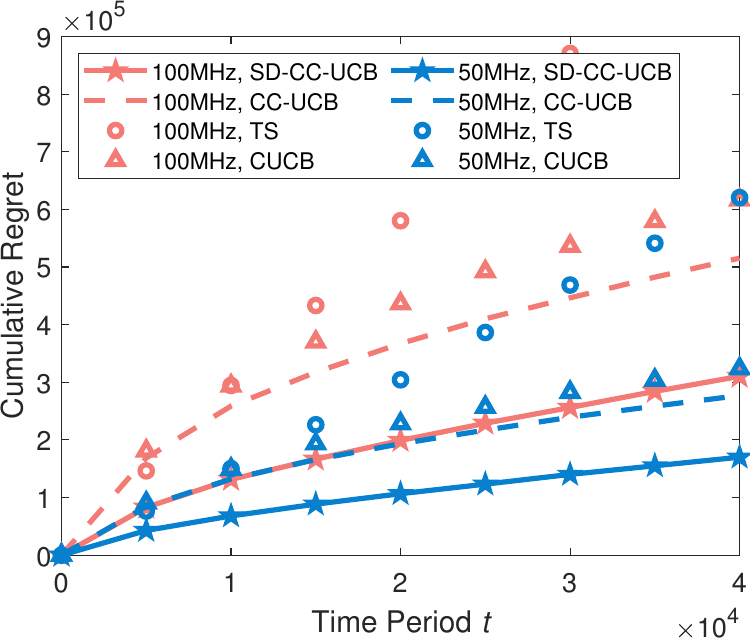}
		\caption{Cumulative regret.}
		\label{Fig_4}
	\end{minipage}
	\begin{minipage}[t]{0.33\textwidth}
		\includegraphics[width=6.3cm]{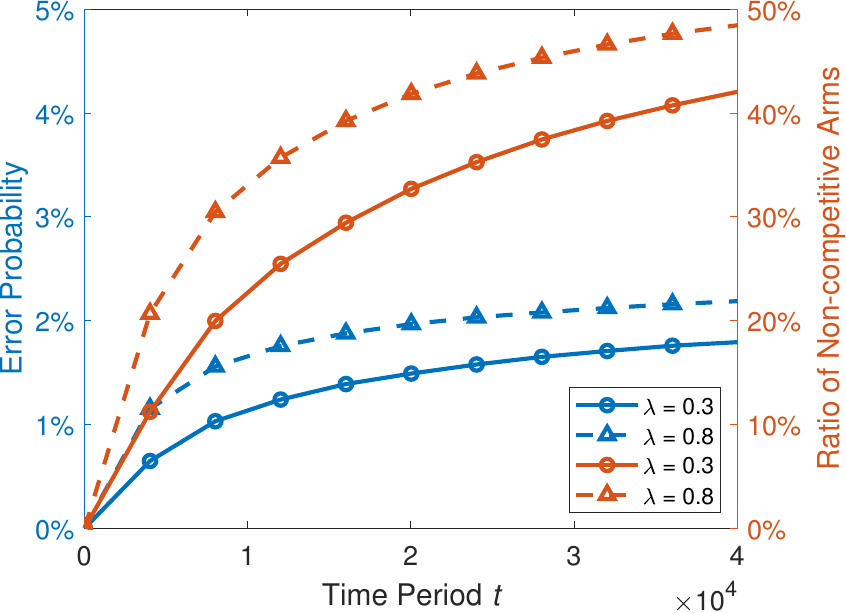}
		\caption{Error probability of CC-UCB identifying non-competitive arms and the ratio of non-competitive arms to all arms.}
		\label{Fig_3}
	\end{minipage}
	\hspace{3mm}
	\begin{minipage}[t]{0.33\textwidth}
		\includegraphics[width=5.7cm]{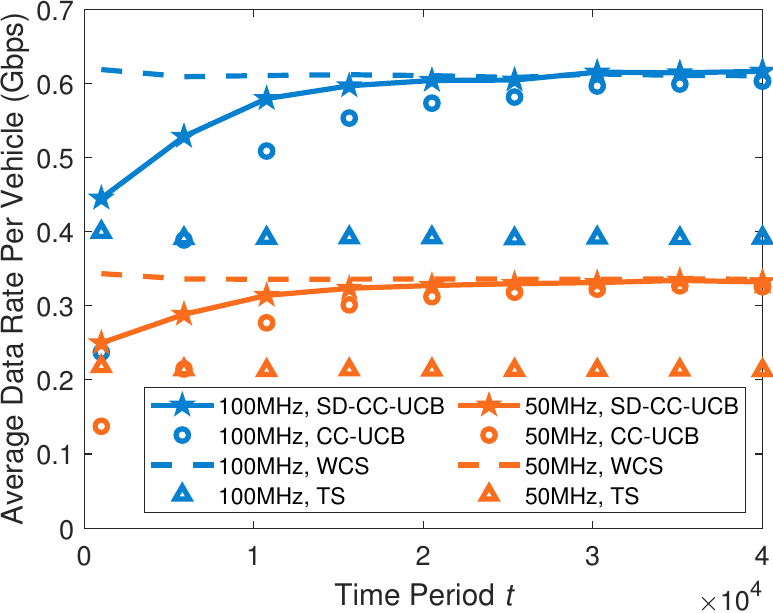}
		\caption{{Average transmission rate per vehicle.}}
		\label{Fig_5}
	\end{minipage}
	
	\begin{minipage}[t]{0.33\textwidth}
		\includegraphics[width=5.9cm]{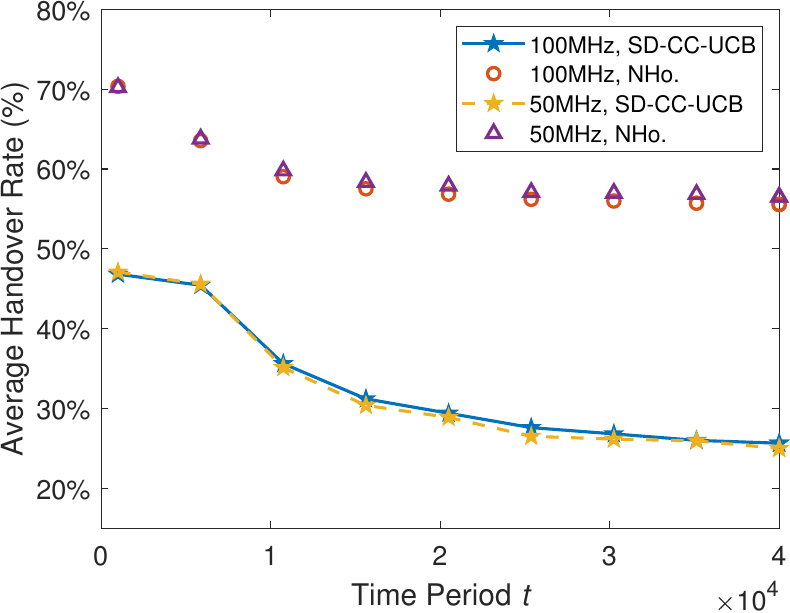}
		\caption{Average handover rate per vehicle.}
		\label{Fig_6}
	\end{minipage}
	\begin{minipage}[t]{0.33\textwidth}
		\includegraphics[width=6.2cm]{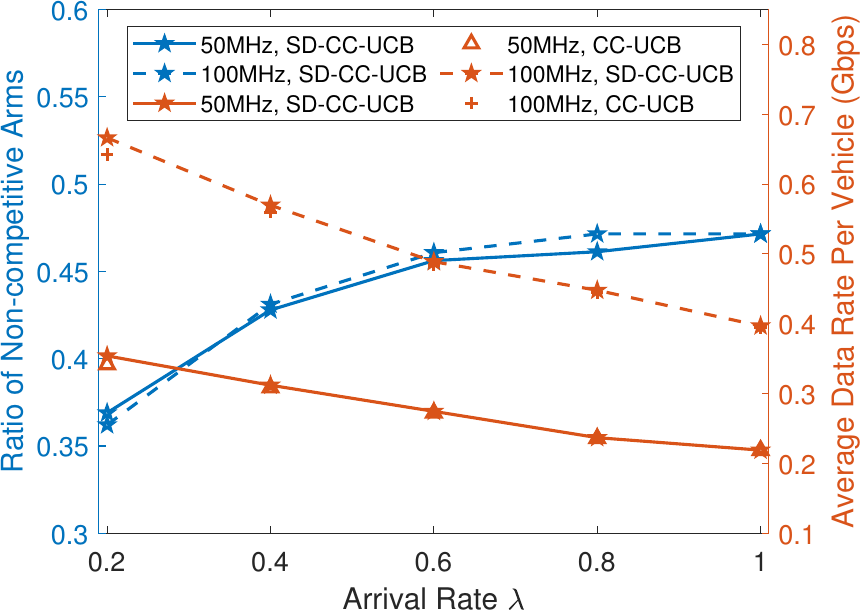}
		\caption{{Ratio of non-competitive arms to all arms and average transmission rate per vehicle against the arrival rate of vehicles.}}
		\label{Fig_7}
	\end{minipage}
	\hspace{3mm}
	\begin{minipage}[t]{0.33\textwidth}
		\includegraphics[width=6.2cm]{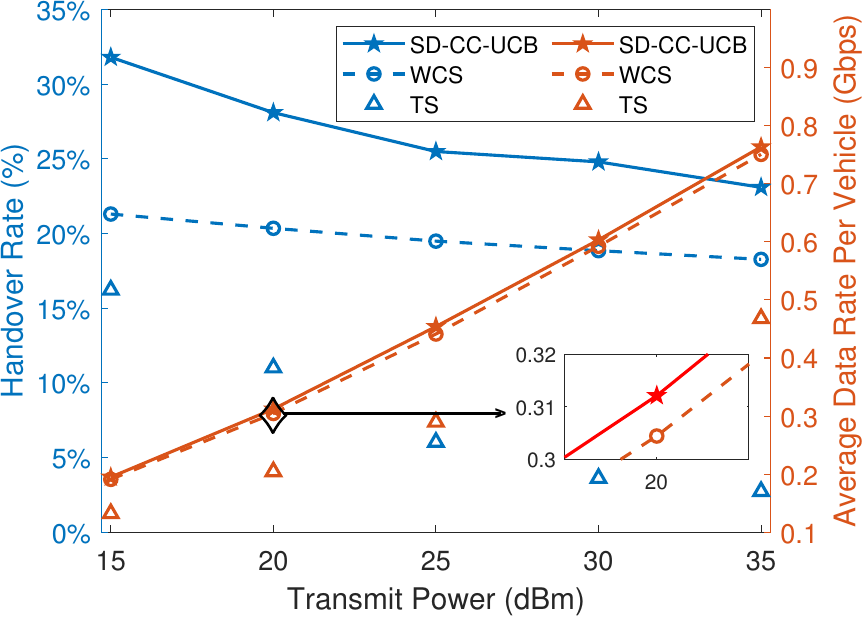}
		\caption{Handover rate and average transmission rate per vehicle with respect to the transmit power.
		}
		\label{Fig_8}
	\end{minipage}
\end{figure*}

Fig.~\ref{Fig_4} shows the cumulative regret against the learning step {given the arrival rate of 0.3.}
The cumulative regret of the SD-CC-UCB algorithm increases mildly with the learning step.
When the period is 20000 and bandwidth is 50MHz, the cumulative regret of the TS algorithm is approximately three times higher than that of the SD-CC-UCB algorithm, and the gap doubles after 40000 periods. 
The fast-fading vehicular channel between the vehicle and BS is non-stationary, violating the assumption of stationary distributions of rewards in the TS algorithm and resulting in divergent regret.
The cumulative regret of CUCB is approximately {20\%} higher than that of CC-UCB after 20000 periods with different bandwidths. 
This difference is attributed to the effective leverage of the correlation of reward distributions on different contexts in CC-UCB, which curtails unnecessary explorations.
Moreover, CC-UCB achieves a regret {1.8} times of SD-CC-UCB across the period. 
In CC-UCB, the BS with the best channel condition is more likely to be selected, then becomes a hot spot, and suffers from significant interference.
However, the SD-CC-UCB algorithm employs TS to detect the variation of the transmission rate caused by interference timely, effectively avoiding the hot spot and mitigating interference.

With a bandwidth of 100MHz, Fig.~\ref{Fig_3} shows the capability of CC-UCB correctly to identify the non-competitive arms, which is critical to the convergence and cumulative regret.
The ratio of identified non-competitive arms continually increases and exceeds {45\%} after 32000 periods when the arrival rate of vehicles is 0.3. 
Ruling out a great portion of non-competitive arms, CC-UCB can eliminate unnecessary explorations to accelerate the convergence, especially in dense vehicular networks with a high arrival rate of vehicles.
When the arrival rate of vehicles is {0.8} and the period reaches 20000, the ratio of non-competitive arms is about {9\%} higher than that with the arrival rate of 0.3. 
As the arrival rate of vehicles increases, the network accommodates more vehicles and updates rewards more frequently, so CC-UCB can identify non-competitive arms more accurately and responsively.
Meanwhile, the probability of incorrectly identifying a competitive arm as a non-competitive arm is below {2.2\%} across the period for different arrival rates of vehicles. 
This result validates that the CC-UCB algorithm successfully explores fewer without sacrificing the reward from exploitation.

Fig.~\ref{Fig_5} presents {the average transmission rate per vehicle} by different algorithms given the bandwidth of 50MHz and 100MHz.
The total transmission rate of the SD-CC-UCB algorithm is approximately {97\%-101\%} compared to that of the WCS algorithm after 16000 periods with different bandwidths. 
The WCS algorithm requires complete CSI of all channels between vehicles and BSs and is invoked at each period, to achieve a near optimal performance.
However, our proposed SD-CC-UCB algorithm works without CSI and updates the reward only based on the transmission rate of the associated BS, achieving comparable performance to the WCS algorithm.

Fig.~\ref{Fig_6} confirms the effectiveness of the SD-CC-UCB algorithm subject to handover cost.
Here, NHo. indicates the results that the SD-CC-UCB algorithm ignores the handover cost.
{With a bandwidth of 50MHz, the handover rate of the SD-CC-UCB algorithm is around 47\% at period 1000, but it decreases to 30\% after 16000 periods.} 
During the initial 16000 periods, handovers occur frequently because the algorithm has not fully learned about the environment, leading to fluctuations in the decision-making process.
As the learning process advances, the agent becomes more acquainted with the environment, leading to a decrease in the handover rate.
After 16000 periods, our proposed algorithm achieves a reduction of approximately {30\%} in the average handover rate compared to the algorithm without considering the handover cost in the case of both bandwidths.
The result demonstrates the effectiveness of incorporating the handover cost into the reward of the TS algorithm. 

In Fig.~\ref{Fig_7} and Fig.~\ref{Fig_8}, we present the performance of the network given the different arrival rates of vehicles and different transmit power.
With a bandwidth of 100MHz, when the arrival rate increases from 0.2 to 1, the average data rate per vehicle decreases by {56\%} as shown in Fig.~\ref{Fig_7}. 
The number of vehicles traveling across the network and the network data load grows with the arrival rate, leading to more severe interference.
{When the arrival rate is 0.2, the gaps in the average data rate per vehicle between CC-UCB and SD-CC-UCB are 12.4Mbps and 23.8Mbps for bandwidths of 50MHz and 100Hz, respectively. 
	CC-UCB cannot manage the traffic overload and compensate the bias caused by interference during the estimation, leading to a poor performance compared to SD-CC-UCB.
	When the arrival rate of vehicles grows to 1, the corresponding gaps decrease to 0.5Mbps and 1.9Mbps. 
	In case of significant traffic congestion, the bias in the reward of CC-UCB is high due to substantial interference.}
{As the arrival rate of vehicles increases from 0.2 to 1, the ratio of non-competitive arms to all arms of SD-CC-UCB with the bandwidth of 100MHz also increases from 36.2\% to 47.1\%, because more vehicles introduces more samples for SD-CC-UCB to identify the non-competitive arms.}

Fig.~\ref{Fig_8} represents the handover rate and the average transmission rate per vehicle against the transmit power.
The bandwidth and arrival rate of vehicles are 100MHz and 0.3, respectively.
When the transmit power increases from 15dBm to 35dBm, the average data rate per vehicle grows from {195}Mbps to {763}Mbps in the SD-CC-UCB algorithm. 
When the transmit power is 15dBm, the handover rate of SD-CC-UCB is 10\% higher than that of WCS as shown in Fig.~\ref{Fig_8}.
However, {the gap decreases to 4.7\% as the transmit power reaches 35dBm}.
Utilizing CSI, the centralized WCS algorithm can predict the transmission rate more accurately compared to SD-CC-UCB based solely on the empirical reward.
At a low transmit power, the SD-CC-UCB probably chooses a sub-optimal BS due to the inaccuracy in learning the transmission rate.
With increased transmit power, the received powers of desired signals are more distinctive, so our proposed algorithm is more likely to identify the optimal BS correctly.
Meanwhile, the SD-CC-UCB algorithm explores less frequently, so the handover rate drops.
More importantly, the results demonstrate that our algorithm achieves the transmission rate within {100\%-103\%} of the WCS algorithm across different transmit power levels. 
Specifically, when the transmit power is 20dBm, the average transmission rate per vehicle achieved by SD-CC-UCB outperforms the WCS algorithm by {8}Mbps. 

\section{Conclusion}
In this paper, we focus on user association in mmWave vehicular networks.
A CC-UCB algorithm is proposed to capture the channel conditions of the fast-fading mmWave vehicular channels correctly and timely by learning the transmission rates with locations and velocities and leveraging the correlations of reward distributions on different contexts.
To further address the impact of interference, handover cost, and other unpredictable and unmeasurable factors on the instantaneous transmission rate, the TS algorithm is employed on each vehicle to narrow the gap between the transmission rate estimated with CC-UCB and the ground truth of the instantaneous rate.
We hope our study inspires more thorough research on this line.




%
\bibliographystyle{IEEEtran}
\bibliography{reference}

\begin{IEEEbiography}[{\includegraphics[width=1in,height=1.25in,clip,keepaspectratio]{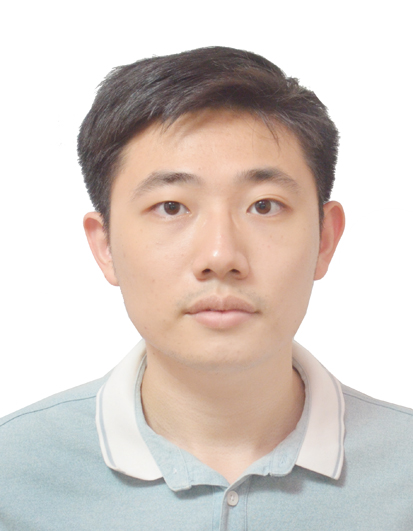}}]{Xiaoyang He}
received the B.S. degree in Communication Engineering from the School of Electronics and Communication Engineering, Sun Yat-sen University, Guangzhou, China, in 2021. He is currently pursuing the Ph.D. degree with Sun Yat-sen University, Shenzhen, China. His research interests include Millimeter-Wave wireless communication channel, Vehicular networks and online learning.
\end{IEEEbiography}

\begin{IEEEbiography}[{\includegraphics[width=1in,height=1.25in,clip,keepaspectratio]{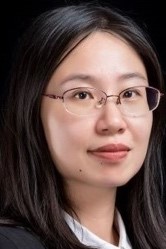}}]{Xiaoxia Huang }
(M' 07) received the B.E. and M.E degrees in electrical engineering both from Huazhong University of Science and Technology, Wuhan, China, in 2000 and 2002, respectively, and the Ph.D. degree in electrical and computer engineering from the University of Florida, Gainesville, FL, USA, in 2007. She is currently a Professor with School of Electronics and Communication Engineering at Sun Yat-Sen University. Her research interests include green wireless communication networks, intelligent wireless networks, and wireless computing.
\end{IEEEbiography}

\begin{IEEEbiography}[{\includegraphics[width=1in,height=1.25in,clip,keepaspectratio]{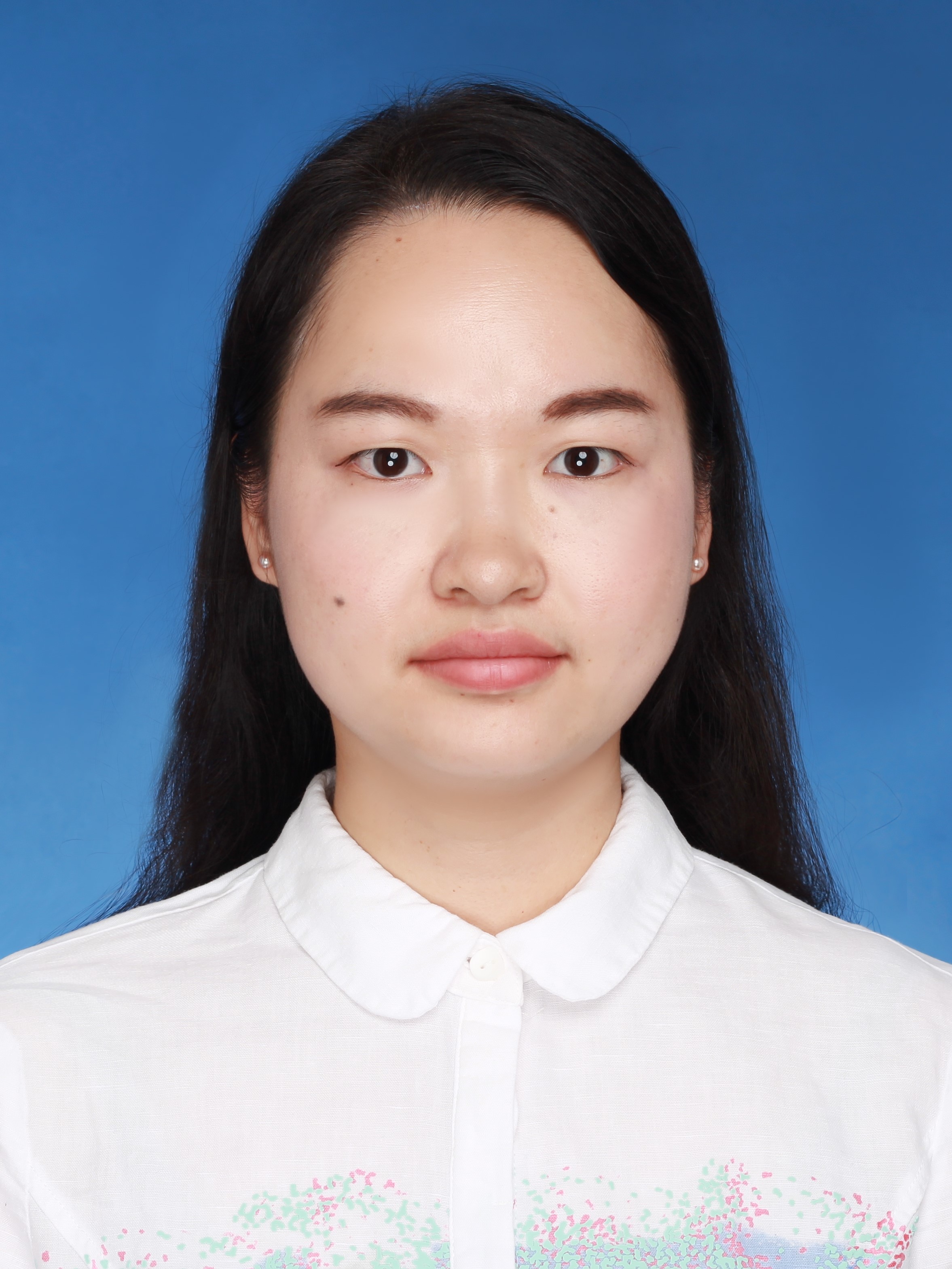}}]{Lanhua Li}
received the B.E. degree and M.E. degree from Hunan Normal University, Changsha, China, in 2013 and 2016, respectively, and the Ph.D. degree in computer applications technology from the University of Chinese Academy of Sciences, ShenZhen, China, in 2021. Since October 2021, she has been a Post-Doctoral Researcher with School of Electronics and Communication Engineering at Sun Yat-Sen University. Her research interests include green wireless communication with focus on resource allocation optimization and energy-efficient multiple access.
\end{IEEEbiography}

\end{document}